%% file: lawm.tex
\documentclass{article}

\usepackage[preprint]{corl_2026} 

\usepackage{graphics}
\usepackage{graphicx}
\usepackage{epsfig}
\usepackage{mathptmx}
\usepackage{times}
\usepackage{amsmath}
\usepackage{amssymb}
\usepackage{float}
\usepackage{placeins}
\usepackage{siunitx}
\usepackage{booktabs}
\usepackage{xcolor}
\usepackage{subfiles}
\usepackage{pgfplots}
\pgfplotsset{compat=1.18}
\usepgfplotslibrary{groupplots}

\usepackage{silence}
\usepackage[caption=false,font=footnotesize]{subfig}

\title{Latent Action Pretraining Through World Modeling}

\author{
\begin{tabular}{c}
\textbf{Bahey Tharwat}$^{1}$ \quad
\textbf{Yara Nasser}$^{2}$ \quad
\textbf{Ali Abouzeid}$^{1}$ \quad
\textbf{Ian Reid}$^{1}$ \\[2pt]
$^{1}$Mohamed bin Zayed University of Artificial Intelligence, Abu Dhabi, UAE \\[1pt]
$^{2}$Alexandria University, Alexandria, Egypt
\end{tabular}
}

\begin{document}
\maketitle

\begin{figure}[!ht]
    \centering
    \includegraphics[width=\linewidth]{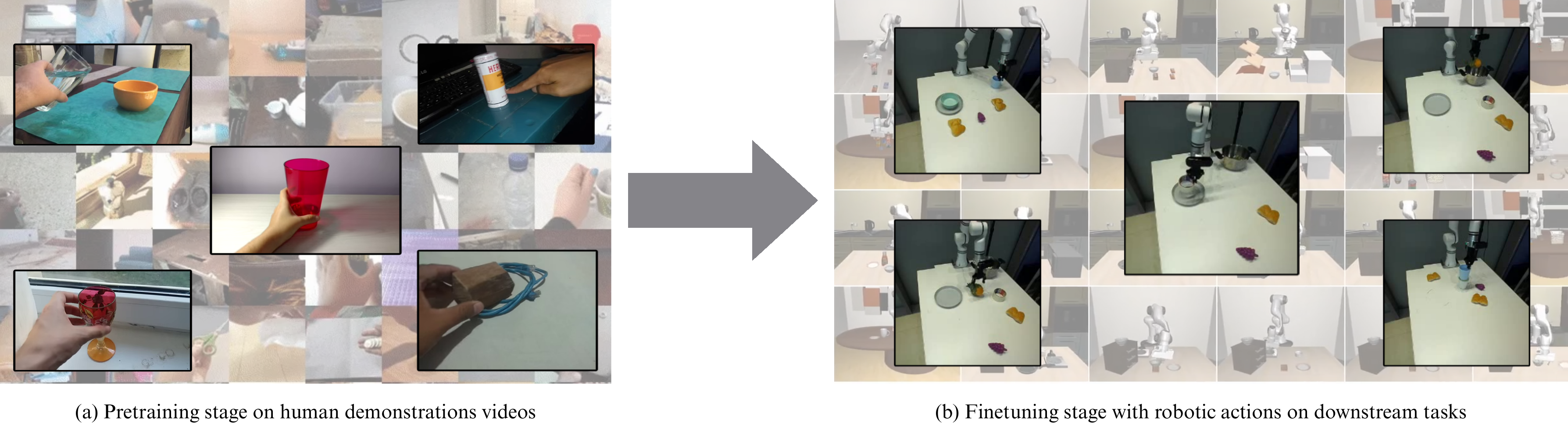}

    \caption{From human demonstrations to robot actions. Our framework uses human demonstration videos for pretraining, then finetunes the pretrained model on robot actions for downstream tasks.}

    \label{fig:first-page}
\end{figure}

\begin{abstract}
Vision-Language-Action (VLA) models have gained popularity for learning robotic manipulation tasks that follow language instructions. State-of-the-art VLAs, such as OpenVLA and $\pi_{0}$, were trained on large-scale, manually labeled action datasets collected through teleoperation. More recent approaches, including LAPA and villa-X, introduce latent action representations that enable unsupervised pretraining on unlabeled datasets by modeling abstract visual changes between frames. Although these methods have shown strong results, their large model sizes make deployment in real-world settings challenging.
In this work, we propose LAWM, a model-agnostic framework to pretrain imitation learning models in a self-supervised way, by learning latent action representations from unlabeled video data through world modeling. These videos can be sourced from robot recordings or videos of humans performing actions with everyday objects. Our framework is able to transfer learned knowledge across tasks, environments, and embodiments. It outperforms models pretrained with ground-truth robot actions and other similar pretraining methods on the LIBERO benchmark and real-world setup, while being efficient and practical for real-world settings.
\end{abstract}


\section{Introduction}

Self-supervised learning has been a key enabler of recent breakthroughs in Large Language Models (LLMs) such as ChatGPT~\citep{chatgpt} and Gemini~\citep{gemini}, where models learn from large amounts of text on the Internet. Inspired by this success, the robotics community is now ready for its own transformative moment, where we can build systems that learn action representations directly from raw, unstructured video data, rather than relying on curated action labels.

Most current approaches to robot learning are heavily based on supervised learning frameworks. Methods like imitation learning and VLA models, including OpenVLA~\citep{OpenVLA} and $\pi_{0}$~\citep{PI0}, require paired image action datasets often obtained through teleoperation. These action annotations are expensive to collect, difficult to scale, and prone to bias, limiting the generalizability of these systems across tasks, environments, and embodiments.

\begin{figure*}[t]
    \centering
    \includegraphics[width=1.0\linewidth]{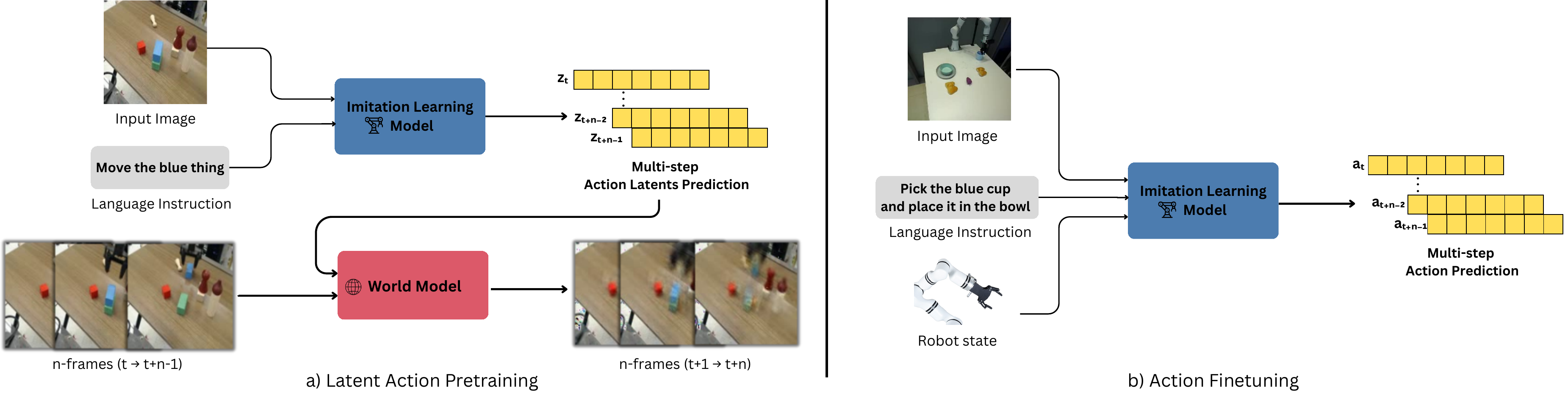}
    \caption{Overview of \textbf{LAWM}, which consists of two stages: 
    (a) \textbf{Latent Action Pretraining}, where an imitation learning model predicts latent actions $z_{t:t+n-1}$ from input images and language instructions. These latents are paired with video frames and jointly optimized with a world model through next-frame prediction.
    (b) \textbf{Action Finetuning}, where the pretrained imitation learning model is adapted to downstream robotic tasks using labeled demonstrations, mapping observations (images, language instructions, and robot state) to ground-truth actions $a_{t:t+n-1}$.}

    \label{fig:VLA-WM}
\end{figure*}

In this work, we introduce \textbf{LAWM}, a \textbf{L}atent \textbf{A}ction pretraining framework through \textbf{W}orld \textbf{M}odeling that aims to overcome these limitations by combining an Imitation Learning Model with a World Model. Our objective, as shown in Figure~\ref{fig:first-page}, is to learn action representations from both robot-collected and human demonstration videos in a fully self-supervised way. These learned representations serve as action priors that can be effectively leveraged during finetuning on downstream tasks. The proposed framework, illustrated in Figure~\ref{fig:VLA-WM}, is designed to be \textit{model-agnostic}, meaning that it does not depend on any specific architecture for the imitation learning model or the world model. This flexibility allows for the integration of a variety of different models.

Our pipeline follows a two-stage approach. The first stage performs end-to-end self-supervised pretraining, where the learning signal comes from next-frame prediction in video sequences. Given an image frame from a human or robot manipulation task and a language instruction, the imitation learning model predicts \textit{action chunk representations}. These $n$ latent actions are used to roll out the world model sequentially from the current frame, predicting each future frame step by step. The second stage finetunes only the imitation learning model on labeled downstream data, while the world model is discarded. As a result, the imitation learning model benefits from a robust prior learned from large-scale unlabeled videos and can be adapted efficiently to downstream robot tasks.

We summarize our main contributions and findings below:
\begin{itemize}
    \item We propose \textbf{LAWM}, a \textit{model-agnostic} framework, to learn action chunk representations for imitation learning models from both robot and human videos without action labels.
    \item Our experiments show that our framework can learn superior action priors from human demonstrations and robotic manipulation videos without using ground-truth action labels.
    \item We demonstrate that our framework with small models such as Diffusion Policy~\citep{diffusionpolicy} and DreamerV3~\citep{DREAMERv3} outperforms methods such as villa-X~\citep{VILLAX} on the LIBERO benchmark~\citep{LIBERO}.
\end{itemize}

\section{Related Work}

Recent advances in Vision-Language Models (VLMs) trained on large-scale internet text, image, and video datasets have paved the way in Robotics for Vision-Language-Navigation (VLN) models~\citep{navgpt, navgpt2, navgpt-mbzuai, navsurvey} and for Vision-Language-Action (VLA) models~\citep{RT2, PI0, OpenXEmbodiment, OpenVLA}. These models extend VLMs by grounding multimodal reasoning in robotic action spaces, enabling robots to follow natural language instructions and generalize to new tasks.
Datasets such as Open X-Embodiment~\citep{OpenXEmbodiment} advance this direction by unifying multi-robot datasets, which supports large-scale models like OpenVLA~\citep{OpenVLA}.

\subsection{Imitation Learning}

Imitation learning methods leverage demonstrations to directly map observations to actions, enabling robots to learn from expert behavior. At the large-scale end of the spectrum, \textbf{OpenVLA}~\citep{OpenVLA} is a 7B parameter VLA model with dual visual encoders (DINOv2~\citep{dinov2} and SigLIP~\citep{siglip}). It is trained on the Open X-Embodiment~\citep{OpenXEmbodiment} dataset with 970k real robot demonstrations. Similarly, \textbf{$\pi_{0}$}~\citep{PI0} is a 3.3B parameter VLA model that leverages a pretrained VLM backbone to inherit Internet-scale knowledge. Taking a contrasting lightweight approach, \textbf{BAKU}~\citep{BAKU} is a compact 7M parameter model that fuses multimodal inputs: vision, proprioception, and task instructions through a FiLM-conditioned encoder~\citep{film} and MLPs. Its observation trunk can be either an MLP or a causal transformer decoder to output actions. Another line of work explores action generation through generative modeling. \textbf{Diffusion Policy}~\citep{diffusionpolicy} frames robot control as a conditional denoising diffusion process that refines noisy action sequences into smooth trajectories. Building on temporal reasoning, \textbf{Action Chunking Transformer (ACT)}~\citep{ACT} tackles long-horizon planning by predicting 'chunks' of future actions instead of single steps, reducing error accumulation and improving efficiency in long tasks.

\subsection{World Models}

World models learn representations of environment dynamics to support planning and control in embodied agents. \textbf{DIAMOND}~\citep{DIAMOND} is a diffusion-based world model that shows strong performance in gaming environments such as CSGO. In contrast, \textbf{Dreamer} world models family~\citep{DREAMERv1, DREAMERv2, DREAMERv3} implements a more sophisticated approach through its \textbf{Recurrent State-Space Model (RSSM)} architecture, which encodes input states into stochastic representations and uses a recurrent sequence model to predict future states given past actions. This RSSM framework enables the agent to learn a world model of its environment and leverage it to plan by \emph{imagining} potential future trajectories, facilitating efficient planning and policy learning in high-dimensional control tasks. Dreamerv3 offers flexible scalability with model sizes ranging from \textbf{1 million to 200 million parameters}.

\subsection{Latent Action Methods}

Early works explored the pretraining of vision encoders~\citep{R3M} on egocentric human videos~\citep{Ego4D}, retargeting human motions to robots~\citep{mimicplay}, or predicting actions from inverse dynamics models~\citep{du2023idm}. Latent action models in simulation and video games~\citep{genie, lapo} showed that abstract action spaces improve generalization, but remained limited to non-robotic domains. Building on these foundations, \textbf{LAPA}~\citep{LAPA} introduced a 7B-parameter Vision-Language-Action model that learns latent actions through a three-stage pipeline, enabling training from unlabeled videos. However, it suffers from framework complexity, inference latency, and high computational cost. \textbf{UniVLA}~\citep{univla} proposes learning cross-embodiment vision-language-action policies by deriving task-centric latent action representations from unlabeled videos. This method addresses task-irrelevant visual noise by decoupling task-relevant dynamics through a two-stage training approach that leverages DINOv2 features and language conditioning. While UniVLA achieves state-of-the-art results with improved computational efficiency, it still relies on relatively large models (7B-parameter Prismatic VLM backbone). More recently, \textbf{Villa-X}~\citep{VILLAX} improves upon LAPA by integrating a proprioceptive forward-dynamics model (FDM) into its Latent Action Model (LAM). This auxiliary component predicts future robot joint angles and gripper states from latent tokens. villa-X demonstrated better performance than LAPA on the SIMPLER benchmark~\citep{simpler}, though it still requires the latent action module during finetuning and remains computationally expensive.

\section{Method}

LAWM consists of two stages: Latent Action Pretraining and Action Fine-tuning.

\subsection{Latent Action Pretraining}

The pretraining stage, illustrated in Figure~\ref{fig:VLA-WM}, is \textit{model-agnostic} and trained end-to-end in a self-supervised way. This allows flexible integration of imitation learning and world model architectures.

Given a current image frame $x_t$ and a language instruction $c$, the imitation learning model $\mathrm{IL}_{\theta}$ predicts a sequence of $n$ latent actions:

\begin{equation}
z_{t:t+n-1} = \mathrm{IL}_{\theta}(x_t, c),
\end{equation}

where $z_{t+n-1} = {z_t, z_{t+1}, \ldots, z_{t+n-1}}$ denotes the predicted latent action sequence. These latent actions are not supervised by ground-truth robot actions. Instead, they are used as inputs to the world model. Specifically, the world model $\mathrm{WM}_{\phi}$ is initialized from the current frame $x_t$ and rolled out sequentially with the $n$ predicted latent actions to predict the next $n$ frames:

\begin{equation}
\hat{x}_{t+1:t+n}
=
\mathrm{WM}_{\phi}
\left(
x_{t:t+n-1},
z_{t:t+n-1}
\right).
\end{equation}

Thus, the model is trained by predicting the shifted future frames $x_{t+1+n}$ step by step from the current frame and the latent actions predicted by $\mathrm{IL}_{\theta}$.

In our implementation, $\mathrm{WM}_{\phi}$ is instantiated as a recurrent state-space model (RSSM), following DreamerV3~\citep{DREAMERv3}. At each rollout step $k \in \{1, \ldots, n\}$, the model updates a deterministic recurrent state $h_{t+k}$ using the previous state $\hat{s}_{t+k-1}$ and latent action $z_{t+k-1}$. It then predicts a stochastic latent state and reconstructs the next frame:

\begin{align}
h_{t+k}
&=
f_{\phi}
\left(
\hat{s}_{t+k-1},
z_{t+k-1}
\right),
\\
u_{t+k}
&\sim
q_{\phi}
\left(
u_{t+k}
\mid
h_{t+k},
x_{t+k}
\right),
\\
\hat{u}_{t+k}
&\sim
p_{\phi}
\left(
\hat{u}_{t+k}
\mid
h_{t+k}
\right),
\\
\hat{x}_{t+k}
&\sim
p_{\phi}
\left(
\hat{x}_{t+k}
\mid
\hat{s}_{t+k}
\right).
\end{align}

Here, $\hat{s}_{t+k} = \left(h_{t+k}, \hat{u}_{t+k}\right)$, where $u_{t+k}$ denotes the internal stochastic state of the world model. The posterior $q_{\phi}$ and prior $p_{\phi}$ are modeled as categorical distributions.

The imitation learning model $\mathrm{IL}_{\theta}$ and world model $\mathrm{WM}_{\phi}$ are optimized jointly using a next-frame reconstruction objective. The training signal is the MSE loss between the predicted frames and the ground-truth frames. Following RSSM-based world models, we also include a KL regularization term between posterior and prior latent distributions, weighted by $\beta$. The final objective is:

\begin{equation}
\mathcal{L}(\theta, \phi)
=
\underbrace{
\frac{1}{n}
\sum_{k=1}^{n}
\left\|
x_{t+k}
-
\hat{x}_{t+k}
\right\|_2^2
}_{\text{MSE loss}}
+
\underbrace{
\beta \mathcal{L}_{\mathrm{KL}}(\phi)
}_{\text{KL Regularizer}} .
\end{equation}

This training objective encourages $\mathrm{IL}_{\theta}$ to produce latent actions that are useful for predicting next frames, thereby grounding the learned action representation in the environment dynamics without requiring ground-truth action supervision.

\subsection{Action Finetuning}

Following the self-supervised pretraining stage, we then proceed to a finetuning stage where the training signal is derived from supervised learning with ground-truth actions. In this stage, the world model is no longer used. Instead, the pretrained imitation learning model (in our case, BAKU~\citep{BAKU} or Diffusion Policy~\citep{diffusionpolicy}) is trained on specific downstream tasks to adapt it to the target environment.

The pretrained model benefits from a robust prior obtained in the pretraining stage from large-scale unlabeled video data, which enables efficient finetuning on labeled datasets. Finetuning can be performed either by (i) adding an additional head that decodes the latent action representation into the ground-truth action space, or (ii) reinitializing the final action latent layer to match the dimensionality of the ground-truth action space and training it accordingly. In our implementation, we adopt the second strategy, reinitializing only the last action latent layer during finetuning as LAPA~\citep{LAPA} demonstrated that this works better empirically.

For both BAKU~\citep{BAKU} and Diffusion Policy~\citep{diffusionpolicy}, finetuning adapts the pretrained imitation learning model $\mathrm{IL}_{\theta}$ to predict ground-truth actions from a collected demonstration dataset $\mathcal{D}$. Each observation $o_t = \left(x_{t,\mathrm{view1}}, x_{t,\mathrm{view2}}, c, p_t\right)$ may contain one or two image views, a language instruction $c$, and proprioceptive state $p_t$. Given $o_t$, the model predicts an action sequence $\hat{a}_{t:t+n-1} = \mathrm{IL}_{\theta}(o_t)$. The finetuning objective can be written as:

\begin{equation}
\mathcal{L}_{\mathrm{IL}}(\theta)
=
\frac{1}{n}
\sum_{k=1}^{n}
\underbrace{
\ell_{\mathrm{act}}
\left(
\hat{a}_{t+k-1},
a_{t+k-1}
\right)
}_{\text{Action prediction loss}} .
\end{equation}

Here, $\ell_{\mathrm{act}}$ denotes the supervised action loss used by the chosen imitation learning architecture, such as a likelihood-based loss for BAKU or a denoising loss for Diffusion Policy. Through this stage, the imitation learning model learns to map image-text inputs to explicit actions, aligning its predictions with supervised ground-truth demonstrations.

\section{Experiments}

We focus on answering the following research questions:
\begin{itemize}
\item \textbf{Q1.} Can LAWM learn superior priors compared to pretraining with ground-truth actions?
\item \textbf{Q2.} How does LAWM perform compared to recent similar methods like villa-X?
\item \textbf{Q3.} Do the learned latent actions correlate with ground-truth actions?
\end{itemize}

\subsection{Benchmarks and Environments}
We evaluate LAWM on diverse datasets and environments for both large-scale pretraining and downstream finetuning, including BridgeData v2, a large-scale robot manipulation dataset, Something-Something v2, a human video dataset featuring interactions with everyday objects, LIBERO, a simulation benchmark for robotic manipulation, and a custom real-world robot manipulation setup. These benchmarks cover a broad range of manipulation tasks and evaluation settings. Additional details regarding the datasets, task suites, and environments are provided in the Appendix ~\ref{benchmarks}.

\begin{figure*}[t!]
    \centering
    \includegraphics[width=1.0\linewidth]{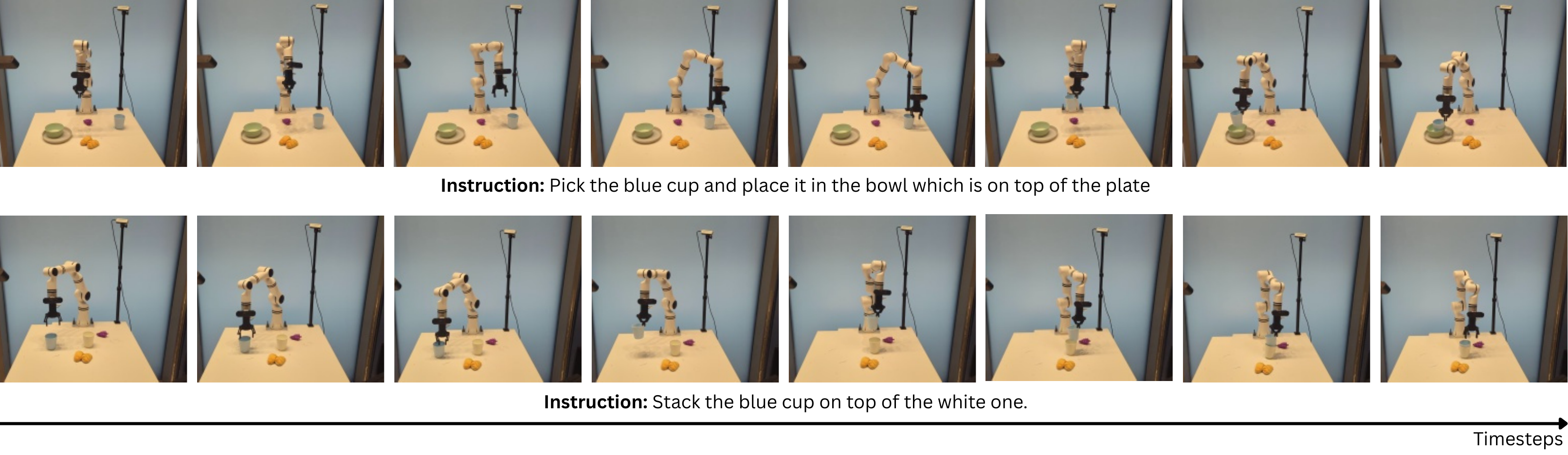}
    \caption{Qualitative results of our framework on the real-world manipulation setup using the 6-DoF Realman robot arm. Top row: Picking the blue cup and placing it in the bowl on top of the plate. Bottom row: Stacking the blue cup on top of the white cup. Images are shown across timesteps.}
    \label{fig:realman-test}
\end{figure*}

\subsection{Baselines and Experimental Setup}

We evaluate LAWM with two imitation learning baselines: BAKU~\citep{BAKU} and Diffusion Policy~\citep{diffusionpolicy}. To address our research questions (\textbf{Q1--Q3}), we design the following experiments (\textbf{E1--E3}):

\textbf{E1. Latent action priors vs. ground-truth pretraining.}  
We establish two performance bounds on the LIBERO-90 benchmark. First, the baseline models are trained from scratch on LIBERO-90. Second, baseline models are pretrained on the ground-truth actions from BridgeData v2~\citep{bridgedata}, then finetuned on LIBERO-90. To evaluate our approach relative to these bounds, we pretrain our framework on each of Something-Something v2~\citep{SomethingSomethingV2} and BridgeData v2~\citep{bridgedata} without action labels, followed by finetuning on LIBERO-90.

\textbf{E2. Comparison with similar methods.}  
We pretrain a Diffusion Policy on human videos from Something-Something v2 dataset, then finetune on the LIBERO task suites (Spatial, Object, Goal, and Long), same as villa-X~\citep{VILLAX} and UniVLA ~\citep{univla}, to enable direct comparison. We also compare with other state-of-the-art models pretrained with supervised learning on ground-truth actions and VLAs that are initialized from VLMs trained on internet scale data.

\textbf{E3. Correlation of latent and ground-truth actions.}  
To quantify the correlation between our learned latent actions and ground-truth actions, we use Canonical Correlation Analysis (CCA). CCA finds linear transformations of two sets of variables such that their representations are maximally correlated. Given latent actions $Z \in \mathbb{R}^{n \times p}$ and ground-truth actions $Y \in \mathbb{R}^{n \times q}$, CCA finds weight vectors $a \in \mathbb{R}^p$ and $b \in \mathbb{R}^q$ that maximize the correlation between transformed variables $u$ and $v$.
\begin{equation}
u = Z a, \quad v = Y b,
\end{equation}

CCA produces pairs of canonical components $(u_i, v_i)$ that capture the linear correlations between latent and ground-truth actions, ordered from strongest to weakest. The first pair $(u_1, v_1)$, known as the first \emph{canonical components} has the strongest correlation. The correlation factor is simply the \textbf{Pearson correlation coefficient} between transformed variables $u$ and $v$.

\begin{equation}
\rho = \frac{\operatorname{cov}(u, v)}{\sqrt{\operatorname{var}(u)} \, \sqrt{\operatorname{var}(v)}}
\end{equation}

Where $\operatorname{cov}(u, v)$ denotes the covariance between $u$ and $v$, and $\operatorname{var}(u), \operatorname{var}(v)$ are their variances.

\section{Results \& Discussion}
We present quantitative and qualitative results of our framework across the LIBERO benchmark, its task suites, and a real-world setup, followed by a discussion of our findings. We use a DreamerV3 world model with a 50M-parameter configuration and a latent action space of dimension 7, matching the dimensionality of the ground-truth actions. The action chunk size is set to 10 future actions for BAKU experiments and 16 for Diffusion Policy experiments. Performance is evaluated using Success Rate (SR), defined as the percentage of successful trials over the total number of attempts. Results are reported across three random seeds, with 30 evaluation trials per task in LIBERO and 10 evaluation trials in the real-world setup. Additional training details are provided in the Appendix ~\ref{training}.

\textbf{R1. Latent action priors vs. ground-truth pretraining.}  
We use BAKU and Diffusion Policy on LIBERO-90 tasks as baselines. An interesting finding is that pretraining on robotics or human video datasets \emph{without} ground-truth actions, when combined with world models, can outperform supervised pretraining on action labels. We hypothesize that supervised pretraining on actions, forces the model to map observations to actions, which can cause it to overfit to the distribution of BridgeData v2 actions (specific robots, trajectories, task setups). However, world model pretraining encourage the model to learn state transitions and dynamics. This generalizes better when finetuned on downstream tasks, since the dynamics are more universal across tasks than a specific embodiment.

\subsubsection*{Results on LIBERO-90}
\begin{figure*}[!t]
    \centering
    \subfloat[Success rates of \textbf{BAKU} on LIBERO-90 under different pretraining strategies.]{
        \includegraphics[width=0.45\linewidth]{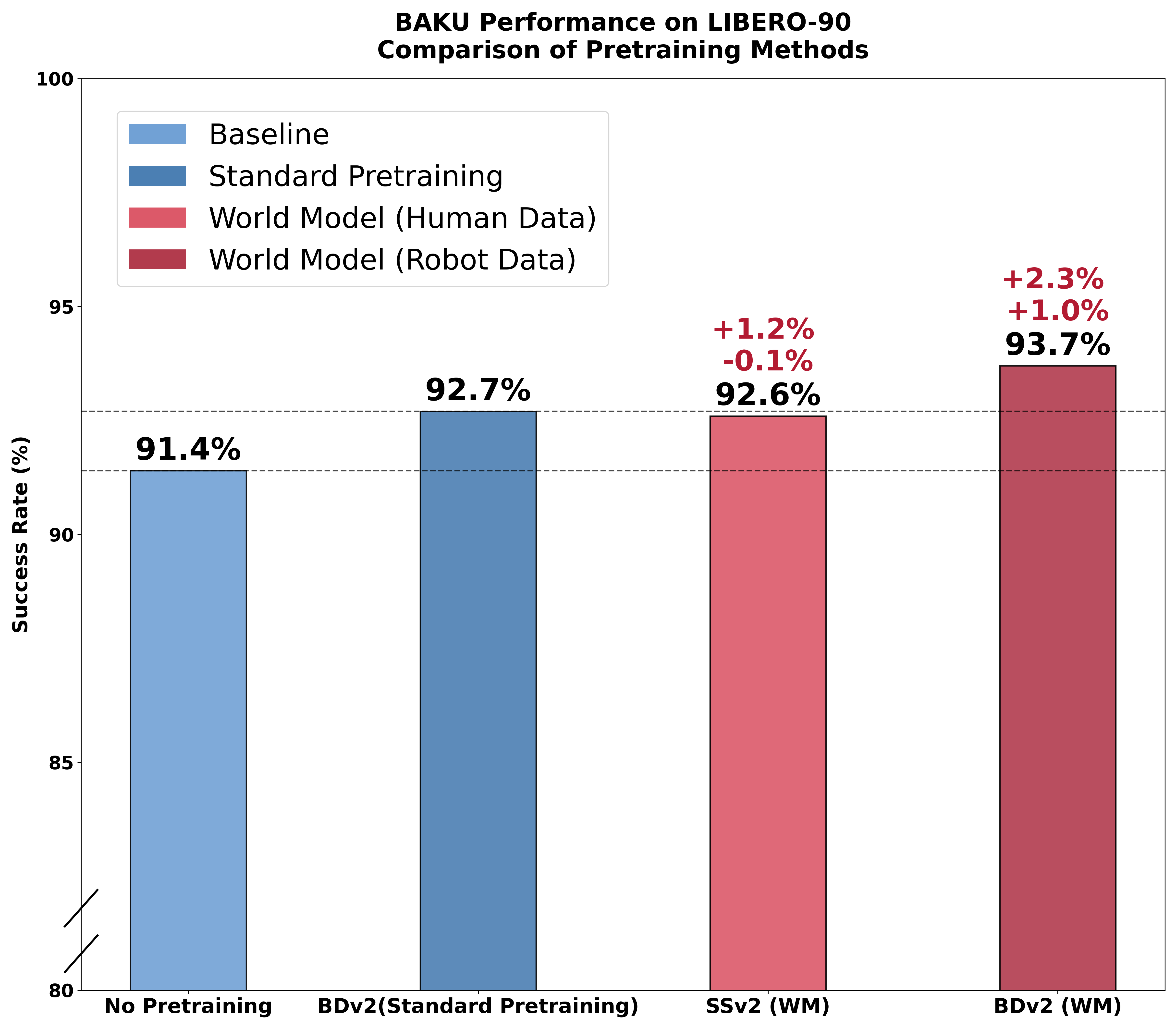}
        \label{fig:baku-bar}
    }
    \hfill
    \subfloat[Success rates of \textbf{Diffusion Policy} on LIBERO-90 under different pretraining strategies.]{
        \includegraphics[width=0.45\linewidth]{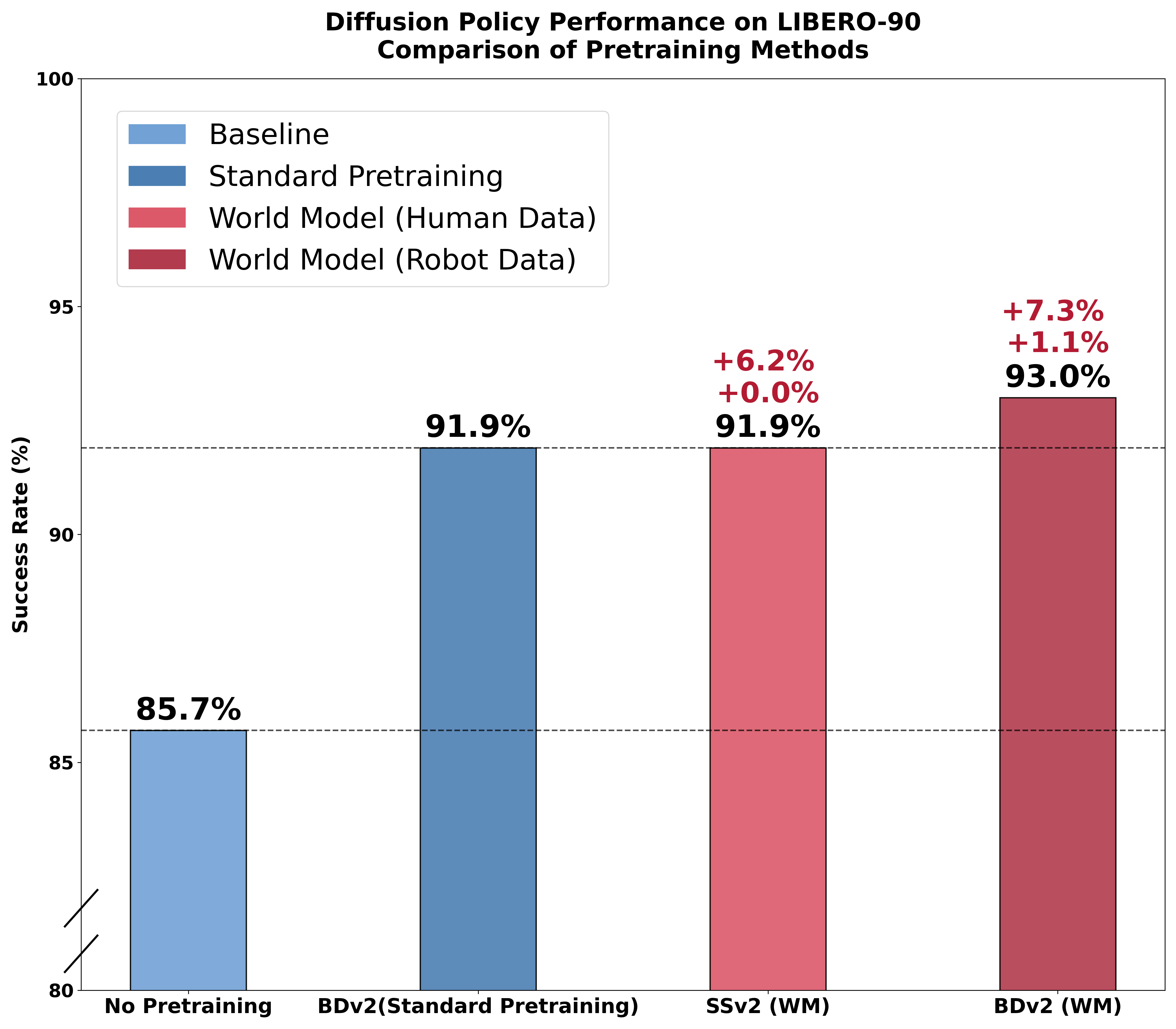}
        \label{fig:dp-bar}
    }
    \caption{Comparison of success rates under different pretraining strategies for (a) BAKU and (b) Diffusion Policy on LIBERO-90 benchmark. WM = world modeling without action labels.}
    \label{fig:baselines-bar}
\end{figure*}

As shown in Figure~\ref{fig:baselines-bar}, Training from scratch provides baselines of 91.4\% (BAKU) and 85.7\% Diffusion Policy, while supervised pretraining on BridgeData improves performance to 92.7\% and 91.9\%, respectively. 
World modeling pretraining results in improvement: BAKU reaches 93.7\% with BridgeData (WM) and 92.6\% with Something-Something (WM), while Diffusion Policy achieves 93.0\% and 91.9\% under the same settings. These results highlight that world modeling not only reduces reliance on action labels but can match or surpass supervised pretraining across models.

\subsubsection*{Results on Real-world setup}
Pretraining our framework solely on human videos from the Something-Something v2 dataset significantly improves performance on real-world tasks, achieving near-perfect success rates across five custom downstream tasks. Training only on our custom dataset without pretraining gives an average success rate of 84\%, which increases to 94\% with our pretraining framework. As shown in Table~\ref{tab:sr-real}, these results show that world modeling improves performance not just in simulation but also in real-world tasks. The qualitative results of our framework on real-world testing are shown in Figure~\ref{fig:realman-test}, demonstrating the ability of our approach to handle complex real-world scenarios.

\begin{table}[H]
\centering
\caption{Success Rates (\%) on five tasks from the real-world manipulation setup.}
\label{tab:sr-real}
\scalebox{0.92}{%
\begin{tabular}{lcccccc}
\toprule
Model & \multicolumn{3}{c}{Pick-and-place} & Move & Stack & Average \\
\cmidrule(lr){2-4}
 & 1 & 2 & 3 &  &  &  \\
\midrule
BAKU w/o latent~\citep{BAKU} & 80.0 & \textbf{100.0} & 90.0 & 70.0 & 80.0 & 84.0 \\
BAKU w/ latent (\textbf{Ours}) & \textbf{90.0} & \textbf{100.0} & \textbf{90.0} & \textbf{90.0} & \textbf{100.0} & \textbf{94.0} \\
\bottomrule
\end{tabular}%
}
\end{table}


\textbf{R2. Comparison with similar methods.}  
We compare our framework using the diffusion policy model with other state-of-the-art models on the LIBERO task suites (Spatial, Object, Goal, Long). Table~\ref{tab:libero-comparison} compares three categories of models:  
1) Models pretrained on ground-truth robotics actions (Octo~\citep{octo}, OpenVLA~\citep{OpenVLA}, $\pi_0$~\citep{PI0}). 2) Models initialized from VLMs trained on internet scale data ($\pi_0$~\citep{PI0} initialized from PaliGemma~\citep{paligemma} and smolVLA~\citep{smolvla} from smolVLM~\citep{smolvlm}).  
3) Models with a similar objective to ours, relying on human videos, robot videos, or a mix of both, such as LAPA~\citep{LAPA}, UniVLA~\citep{univla}, and villa-X~\citep{VILLAX}. We first train the diffusion policy on the LIBERO task suites without any pretraining. Providing a baseline that underperforms UniVLA and villa-X, with our proposed pretraining framework on Something-Something v2 dataset, we further improve upon the diffusion policy, achieving the best average success rate across task suites.

\begin{table}[!t]
\centering
\caption{Baseline models' performance comparison across LIBERO task suites. Reported SR scores are taken directly from the original publications. LAPA results are reproduced by UniVLA authors.}
\label{tab:libero-comparison}
\scalebox{0.85}{%
\begin{tabular}{llccccc}
\toprule
Model & Pretraining Method & Spatial & Object & Goal & Long & Average \\
\midrule
Octo~\citep{octo} & Robot Actions & 78.90 & 85.70 & 84.60 & 51.10 & 75.10 \\
OpenVLA~\citep{OpenVLA} & Robot Actions & 84.70 & 88.40 & 79.20 & 53.70 & 76.50 \\
$\pi_{0}$~\citep{PI0} & Robot Actions & 90.00 & 86.00 & 95.00 & 73.00 & 86.00 \\
\midrule
$\pi_{0}$ (Paligemma-3B)~\citep{PI0} & VLM Checkpoint & 87.00 & 63.00 & 89.00 & 48.00 & 71.80 \\
SmolVLA (SmolVLM-2.25B)~\citep{smolvla} & VLM Checkpoint & 93.00 & 94.00 & 91.00 & 77.00 & 88.75 \\
\midrule
LAPA~\citep{LAPA} & Mix Videos & 73.80 & 74.60 & 58.80 & 55.40 & 65.70 \\
UniVLA~\citep{univla} & Human Videos & 91.20 & 94.20 & 90.20 & \underline{79.40} & 88.70 \\
villa-X~\citep{VILLAX} & Mix Videos & \textbf{97.50} & \textbf{97.00} & \underline{91.50} & 74.50 & \underline{90.10} \\
\midrule
Diffusion Policy w/o latent~\citep{diffusionpolicy} & None & 88.10 & 81.40 & 87.50 & 70.50 & 81.87 \\
Diffusion Policy w/ latent (\textbf{Ours}) & Human Videos & \underline{92.70} & \underline{96.30} & \textbf{94.20} & \textbf{88.00} & \textbf{92.80} \\
\bottomrule
\end{tabular}%
}
\end{table}

\textbf{R3. Correlation of latent and ground-truth actions.}  
We group actions by task type to assess correlation across manipulation tasks. As shown in Table~\ref{tab:cca_first_coeff}, the first canonical component achieves strong correlations in all task categories: Put (0.9154), Move (0.9599), Remove (0.9098), and Take (0.9541). Our method shows better correlation factors than villa-X~\citep{VILLAX} in most tasks, which we attribute to the fact that our latent actions are predicted by an imitation learning model conditioned on task instruction, aligning more closely with ground-truth actions. In contrast, villa-X relies on inverse dynamics models, whose latents primarily capture scene transitions rather than actions.

\begin{table}[H]
\centering
\caption{CCA First Coefficient per group of tasks}
\label{tab:cca_first_coeff}
\scalebox{0.95}{%
\begin{tabular}{lcccc}
\toprule
Method & Put & Move & Remove & Take \\
\midrule
villa-X~\citep{VILLAX} & 0.7863 & 0.9174 & \textbf{0.9955} & 0.9270 \\
LAWM (\textbf{Ours}) & \textbf{0.9154} & \textbf{0.9599} & 0.9098 & \textbf{0.9541} \\
\bottomrule
\end{tabular}%
}
\end{table}

\section{Limitations}
Despite the proposed improvements, this work still has limitations. First, jointly training an imitation learning model and a world model can be computationally expensive, especially with recent VLAs and world models with billions of parameters. Second, training the world model with MSE image reconstruction is known for rewarding pixel accuracy over task-relevant dynamics (blur, background shortcuts). Adding perceptual loss, optical flow loss, or masked modeling could help reduce shortcut learning. Third, the world model is trained only on single-view visual observations, ignoring multi-view inputs, proprioceptive data, and gripper states. This may limit its understanding of manipulation dynamics and scene geometry. Including these sources of information could help the model learn more physically grounded latent actions. We leave these explorations for future work.

\section{Conclusion}
In this paper, we introduce LAWM, a Latent Action pretraining framework for imitation learning models to learn high-quality action representations through World Modeling from unlabeled videos. We demonstrate that our method improves model performance in downstream tasks compared to supervised pretraining in the LIBERO benchmark and real-world setup. We also show that our method can be applied to human demonstration videos, where explicit action information is absent and the embodiment gap is substantial. LAWM provides a scalable way to learn action priors without relying on action labels. We expect that our work will open up the potential for building foundation models for robotics by pretraining large models on much larger web-scale video data.

\clearpage


\bibliography{references}

\clearpage

\subfile{appendix}
\end{document}

%% file: appendix.tex
\newcommand{\appendixtitlepage}{
    \clearpage
    \begin{center}
        {\LARGE \bfseries Appendix\par}
        \vspace{0.5em}
    \end{center}
    \vspace{2em}
}

\ifSubfilesClassLoaded{%
    \title{Appendix \\[0.5em] \large Latent Action Pretraining Through World Modeling}
    \author{}
    \date{}
    \maketitle
}{%
    \appendixtitlepage
}

\appendix

\begin{figure}[H]
\centering
\begin{tikzpicture}
\begin{groupplot}[
    group style={
        group size=3 by 1,
        horizontal sep=0.9cm
    },
    width=0.4\linewidth,
    height=0.4\linewidth,
    grid=both,
    grid style={line width=.1pt, draw=gray!20},
    major grid style={line width=.2pt, draw=gray!35},
    tick label style={font=\scriptsize},
    label style={font=\scriptsize},
    legend style={
        draw=none,
        fill=white,
        font=\scriptsize
    },
    legend cell align={left}
]

\nextgroupplot[
    xlabel={WM size (M)},
    ylabel={Success rate (\%)},
    xmin=0, xmax=105,
    ymin=85, ymax=92,
    xtick={1,12,25,50,100},
    ytick={86,88,90,92},
    legend pos=south east
]
\addplot[
    thick,
    red,
    mark=*,
    smooth
] coordinates {
    (1,86.7)
    (12,87.7)
    (25,90.0)
    (50,91.0)
    (100,91.0)
};
\addlegendentry{BAKU w/ latent}

\addplot[
    thick,
    blue,
    dashed
] coordinates {
    (0,89.0)
    (105,89.0)
};
\addlegendentry{BAKU baseline}

\nextgroupplot[
    xlabel={WM size (M)},
    ylabel={PSNR (dB)},
    xmin=0, xmax=105,
    ymin=17, ymax=24.5,
    xtick={1,12,25,50,100},
    ytick={18,20,22,24},
    legend pos=south east
]
\addplot[
    thick,
    red,
    mark=*,
    smooth
] coordinates {
    (1,17.64638557)
    (12,20.76683859)
    (25,21.80623069)
    (50,23.14769807)
    (100,23.75329798)
};
\addlegendentry{World model PSNR}

\nextgroupplot[
    xlabel={WM PSNR (dB)},
    ylabel={Success rate (\%)},
    xmin=17, xmax=24.5,
    ymin=85, ymax=92,
    xtick={18,20,22,24},
    ytick={86,88,90,92},
    legend pos=south east
]
\addplot[
    thick,
    red,
    mark=*,
    smooth
] coordinates {
    (17.64638557,86.7)
    (20.76683859,87.7)
    (21.80623069,90.0)
    (23.14769807,91.0)
    (23.75329798,91.0)
};
\addlegendentry{BAKU w/ latent}

\addplot[
    thick,
    blue,
    dashed
] coordinates {
    (17,89.0)
    (24.5,89.0)
};
\addlegendentry{BAKU baseline}

\end{groupplot}
\end{tikzpicture}
\caption{Effect of world model size and reconstruction quality on BAKU performance on the LIBERO-Long benchmark. Left: success rate improves as the world model size increases, surpassing the BAKU baseline of 89\%. Middle: larger world models achieve higher PSNR. Right: success rate generally increases with world model PSNR, suggesting that improved world model prediction quality correlates with better downstream policy performance.}
\label{fig:wm_size_psnr_sr_libero10}
\end{figure}

\section{Additional Experiments}

In addition to the main results, we conduct a series of experiments to better understand the properties of action priors learned from large-scale human manipulation videos. Specifically, we investigate (1) how scaling the world model affects downstream performance and reconstruction quality, (2) whether pretraining reduces the amount of expert robot demonstrations required during finetuning, (3) whether pretrained policies solve tasks more efficiently at test-time, and (4) qualitative results of image pairs that exhibit similar latent actions and results of world model rollouts.

\subsection{Scaling World Model Size}

We investigate how the capacity of the pretrained world model affects downstream robotic manipulation performance. Specifically, we train world models with different numbers of parameters on the Something-Something v2 dataset, and then use the learned representations to finetune BAKU on the challenging LIBERO-Long benchmark.

Figure~\ref{fig:wm_size_psnr_sr_libero10} shows that increasing the world model size consistently improves its visual prediction quality. The Peak Signal-to-Noise Ratio (PSNR), which represents the reconstructed image quality, increases from 17.65 for the 1M-parameter model to 23.75 for the 100M-parameter model, indicating that larger world models learn more accurate predictive representations. This trend suggests that world model prediction quality benefits directly from increased model capacity.

More importantly, this improvement in world model quality translates to better downstream finetuning performance. The LIBERO-Long success rate increases from 86.7\% with the 1M world model to 91.0\% with the 50M and 100M models. Smaller world models remain below the BAKU baseline of 89.0\%, whereas larger models surpass it, reaching 90.0\% at 25M parameters and 91.0\% at 50M parameters. This suggests that sufficiently large world models provide more useful representations for downstream policy finetuning.

However, we observe that downstream performance begins to saturate at larger model sizes. Although PSNR continues to improve when scaling the world model from 50M to 100M parameters, the success rate remains fixed at 91.0\%. This indicates that improved visual prediction quality alone may not always yield proportional gains in downstream control once the policy reaches a certain performance level. Consistent with scaling-law behavior~\citep{scalinglaws}, we hypothesize that further gains may require scaling both the world model capacity and the amount of pretraining data. Since our current experiments use only 10\% of Something-Something v2, increasing the data scale together with model size may allow larger world models to learn richer predictive representations and further improve downstream manipulation performance.

\subsection{Reducing Expert Demonstration Requirements}
A key motivation for pretraining is to reduce the amount of expert robot data required for downstream finetuning. To evaluate this, we finetune a policy pretrained on human video data with varying numbers of expert robot demonstrations and compare its performance against a diffusion policy baseline trained without latent pretraining on the LIBERO-Long benchmark.

Starting from the standard setting of 50 expert demonstrations, we evaluate finetuning with 5, 10, 20, 30, 40, and 45 demonstrations, corresponding to 10\%, 20\%, 40\%, 60\%, 80\%, and 90\% of the full demonstration budget. Figure~\ref{fig:libero_long_ft_demos} shows that increasing the number of finetuning demonstrations generally improves the performance of the pretrained policy, with the success rate rising from 46.0\% with 5 demonstrations to 88.0\% with 50 demonstrations.

Importantly, latent pretraining enables effective finetuning with substantially fewer expert demonstrations. With only 20 demonstrations, the pretrained policy achieves a success rate of 77.0\%, outperforming the baseline diffusion policy trained from scratch with the full 50 demonstrations, which achieves 70.5\%. Since 20 demonstrations correspond to only 40\% of the full 50-demonstration setting, this means that the pretrained policy requires less than half of the expert robot demonstrations while still surpassing the fully trained baseline.

These results indicate that the action priors learned during pretraining substantially improve the sample efficiency during finetuning. In particular, latent pretraining reduces the dependence on costly expert robot demonstrations and enables the policy to achieve strong downstream performance even in relatively low-data finetuning regimes.

\begin{figure}[!ht]
\centering
\begin{tikzpicture}
\begin{axis}[
    width=0.72\linewidth,
    height=0.48\linewidth,
    xlabel={Number of expert demonstrations},
    ylabel={Success rate (\%)},
    xmin=0, xmax=55,
    ymin=40, ymax=100,
    xtick={5,10,20,30,40,45,50},
    ytick={40,40,50,60,70,80,90},
    grid=both,
    grid style={line width=.1pt, draw=gray!20},
    major grid style={line width=.2pt, draw=gray!35},
    legend style={
        at={(0.97,0.03)},
        anchor=south east,
        draw=none,
        fill=white,
        font=\small
    },
    tick label style={font=\small},
    label style={font=\small},
]

\addplot[
    thick,
    red,
    mark=*,
] coordinates {
    (5,46.0)
    (10,64.3)
    (20,77.0)
    (30,81.7)
    (40,83.0)
    (45,86.0)
    (50,88.0)
};
\addlegendentry{Diffusion Policy w/ latent}

\addplot[
    thick,
    blue,
    dashed
] coordinates {
    (0,70.5)
    (55,70.5)
};
\addlegendentry{Diffusion Policy Baseline}

\end{axis}
\end{tikzpicture}
\caption{Success rate of diffusion policy on the LIBERO-Long benchmark as a function of the number of expert demonstrations used for finetuning. The dashed horizontal line denotes the baseline trained from scratch on 50 demonstrations (100\%) achieving 70.5\%, while the red line shows that increasing the number of finetuning demonstrations consistently improves performance, achieving 88.0\% with 50 demonstrations.}
\label{fig:libero_long_ft_demos}
\end{figure}

\subsection{Task Completion Efficiency}

Beyond task success, we investigate whether pretraining improves the efficiency with which policies complete tasks. Specifically, we measure the average number of environment steps required to successfully solve tasks after finetuning.

Table~\ref{tab:completion_steps} compares a pretrained and a non-pretrained diffusion policy on the LIBERO task suites. We find that the model initialized from pretrained world models completes tasks in fewer interaction steps on average, requiring an average of \texttt{81} steps compared to \texttt{102} steps for policies trained from scratch, which is a 20\% reduction in completion steps. This result suggests that pretraining not only improves success rates but also leads to more efficient behavior.

\begin{table}[!ht]
\centering
\caption{Number of completion steps across task suites. Lower is better.}
\label{tab:completion_steps}
\scalebox{1.0}{%
\begin{tabular}{lccccc}
\toprule
Method & Spatial & Object & Goal & Long & Average\\
\midrule
Diffusion Policy Baseline & 94 & 62 & 82 & 170 & 102 \\
Diffusion Policy w/ latent \textbf{(Ours)} & \textbf{89} & \textbf{49} & \textbf{54} & \textbf{133} & \textbf{81 }\\
\bottomrule
\end{tabular}%
}
\end{table}

\subsection{Qualitative Results}

\subsubsection{Latent Actions Similarity}
We perform a qualitative analysis of the learned latent actions by computing the cosine similarity between latent action representations, and then visualize the corresponding image pairs with high similarity scores. As shown in Figure~\ref{fig:action_similarities}, these pairs correspond to similar underlying robot behaviors, highlighting that the learned latent action captures a meaningful action-related structure.

\begin{figure}[!ht]
    \centering
    \includegraphics[width=1.0\linewidth]{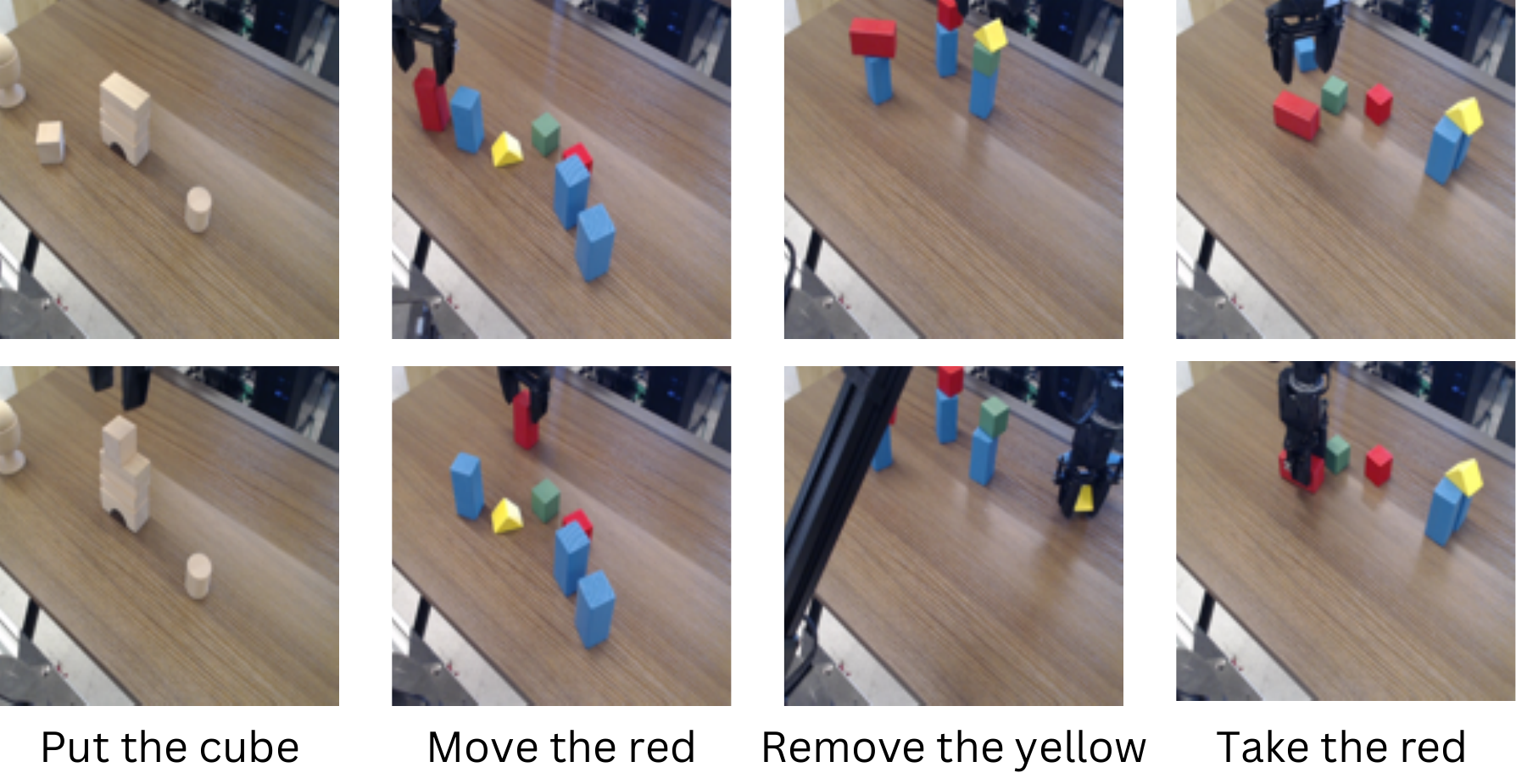}
    \caption{Examples of image pairs exhibiting similar latent actions}
    \label{fig:action_similarities}
\end{figure}

\subsubsection{World Model Rollouts}

We further visualize qualitative examples of world model rollouts after the pretraining stage. Given an initial frame and a predicted action chunk from the imitation learning model, the world model predicts the future frames. We emphasize that high-fidelity video generation is not an objective of our approach, since the world model is not used in any further stages. Instead, these rollouts serve as a qualitative check that the learned latent actions are grounded in meaningful video dynamics. As shown in Figure~\ref{fig:wm_rollouts}, the predicted rollouts capture coherent object motion and scene changes, suggesting that the latent actions encode transitions that can support downstream policy finetuning.

\begin{figure}[!ht]
    \centering
    \includegraphics[width=1.0\linewidth]{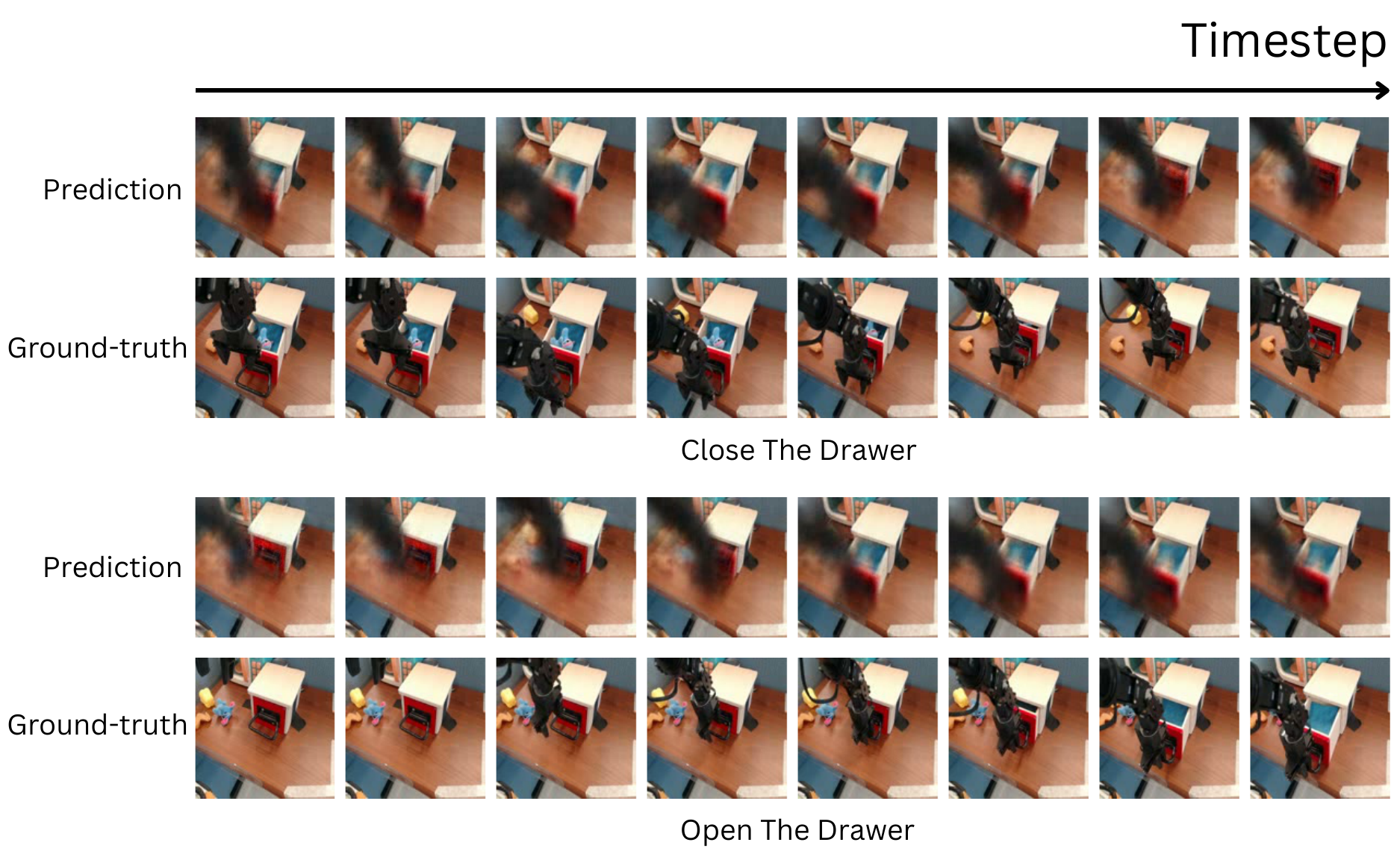}
    \caption{Qualitative examples of world model rollouts using latent actions after pretraining.}
    \label{fig:wm_rollouts}
\end{figure}

\section{Implementation and Training Details}
\label{training}
Our framework consists of an imitation learning model and a world model used only during pretraining. The imitation learning model receives image observations, a language instruction, and a proprioceptive state. Image observations are encoded with a ResNet-18 image encoder. The model supports two camera views, a static third-person view, and a gripper-mounted view, and both views share the same image encoder weights. Language instructions are encoded with a CLIP text encoder to produce a task embedding, while proprioceptive inputs are encoded with a lightweight MLP. The resulting image, language, and proprioceptive embeddings are fused by a fusion network. For BAKU, the fusion module is a GPT-style transformer that receives a sequence of tokens containing the language embedding, visual tokens from each camera view, the proprioceptive token, and a learned action token. The output at the action token is passed to an MLP action head that predicts a chunk of future actions. For Diffusion Policy, the image and proprioceptive embeddings are concatenated and projected into an observation embedding, which is then concatenated with the language embedding and used as global conditioning for a conditional one-dimensional U-Net. The U-Net serves as the action prediction head and denoises a sequence of noisy actions using a DDIM diffusion process.

During pretraining, the framework uses a single image view and the text instruction, while the second camera view and proprioceptive inputs are set to zero. Importantly, the full imitation learning model architecture is still instantiated during pretraining, which keeps the network structure identical between pretraining and finetuning. Since the image encoder is shared across camera views, the encoder trained on the single available pretraining view can be reused for both views during finetuning. The proprioceptive encoder is also present during pretraining but receives zero inputs, so it begins learning meaningful proprioceptive representations during finetuning. The world model, Dreamerv3, which is only used in pretraining, consists of an image encoder that maps images to latent states, a recurrent dynamics model that predicts future latent states conditioned on actions, and a decoder that reconstructs images from latent states. The policy and world model use separate image encoders; sharing these encoders is an interesting direction for future work.

During finetuning, the world model is discarded and only the imitation learning policy is optimized. All imitation learning model components are finetuned, including the image encoder, language projection, proprioceptive encoder, fusion network, and action head. If the latent action dimension used during pretraining differs from the ground-truth robot action dimension, the final action prediction layer must be reinitialized to match the downstream action space. In our robot setup, the ground-truth action is 7-dimensional, corresponding to a 6-DoF end-effector pose and a gripper state. Since we use a 7-dimensional latent action space during pretraining, no action layer reinitialization is required, and the complete pretrained policy is reused as is during finetuning. 

Table~\ref{tab:pretraining_configs} summarizes the hyperparameters used in both the pretraining and finetuning stages. Unless otherwise stated, these settings are shared across all datasets, simulation experiments, and real-world experiments.

\begin{table*}[t]
\centering
\caption{Hyperparameters used for pretraining and finetuning. All values are shared across different datasets, simulation, and
real-world experiments.}
\label{tab:pretraining_configs}

\begin{minipage}[t]{0.46\textwidth}
\centering
\small
\textbf{Pretraining}
\vspace{2mm}

\begin{tabular}{lc}
\toprule
Name & Value \\
\midrule
Epochs & 50 \\
Batch size & 256 \\
Optimizer & AdamW \\
Learning rate & $1 \times 10^{-4}$ \\
Weight decay & $1 \times 10^{-6}$ \\
Gradient clipping & 100 \\
Image resolution & 128 \\
Task embedding dimension & 512 \\
Latent Action dimension & 7 \\
Proprioception dimension & - \\
World model enabled & True \\
Gripper camera enabled & False \\
Proprioception enabled & False \\
\bottomrule
\end{tabular}
\end{minipage}
\hfill
\begin{minipage}[t]{0.46\textwidth}
\centering
\small
\textbf{Finetuning}
\vspace{2mm}

\begin{tabular}{lc}
\toprule
Name & Value \\
\midrule
Epochs & 105 \\
Batch size & 256 \\
Optimizer & AdamW \\
Learning rate & $1 \times 10^{-4}$ \\
Weight decay & $1 \times 10^{-6}$ \\
Gradient clipping & 100 \\
Image resolution & 128 \\
Task embedding dimension & 512 \\
Action dimension & 7 \\
Proprioception dimension & 5 \\
World model enabled & False \\
Gripper camera enabled & True \\
Proprioception enabled & True \\
\bottomrule
\end{tabular}
\end{minipage}

\end{table*}

\section{Benchmarks and Environments Details}
\label{benchmarks}

We evaluate LAWM across diverse datasets and environments that enable both large-scale pretraining and downstream finetuning. An overview of the benchmarks used in our work is shown in Figure~\ref{fig:datasets}.

\begin{itemize}
    \item \textbf{BridgeData v2} (BDv2)~\citep{bridgedata}: A large-scale dataset comprising 60,096 trajectories collected across 24 environments using a low-cost robot. It supports multi-task and multi-environment learning, featuring diverse manipulation tasks such as pick-and-place, pushing, and folding, with over 100 objects involved.

    \item \textbf{Something-Something v2} (SSv2)~\citep{SomethingSomethingV2}: A video dataset containing 220,847 labeled video clips of humans performing actions with everyday objects. Although not a robotic dataset, it is widely used for training models in multi-modal and temporal understanding tasks.

    \item \textbf{LIBERO Benchmark}~\citep{LIBERO}: The LIBERO benchmark contains LIBERO-90, which are 90 tasks spanning diverse objects, layouts, and goals, and is often used as the default evaluation benchmark in many works. LIBERO also has four task suites: LIBERO-Spatial evaluates the model performance in novel layouts with the same tasks and object types, LIBERO-Object evaluates the model performance with novel object types with the same tasks and layouts, LIBERO-Goal evaluates the model performance under novel tasks with the same types and layouts of objects, and LIBERO-Long evaluates the model's performance under a diverse set of objects, layouts, and backgrounds, focusing on long-horizon tasks. Each task suite contains 10 tasks with 50 demonstrations per task for finetuning.

    \item \textbf{Real-world Setup}: We further validate our approach on a real-world tabletop manipulation setup using a 6-DoF Realman robot arm equipped with a 1-DoF gripper. We collect a custom dataset comprising five tasks: three pick-and-place tasks, one stacking task, and one object-moving task. Each scene was captured from two fixed, distinct camera viewpoints to provide multi-view observations. For each task, we recorded 50 demonstrations using human teleoperation with a Meta Quest VR controller. The task instructions are:
    
    \textbf{Pick-and-Place:} (i) Put both cups on the plate, (ii) Pick the blue cup and place it in the bowl which is on top of the plate, (iii) Pick the pineapple out of the pot and place it in between the cups; \textbf{Move:} Move the pineapple to the pot and the roll to the plate; \textbf{Stack:} Stack the blue cup on top of the white one.
\end{itemize}

\begin{figure}[!ht]
    \centering
    \includegraphics[width=1.0\linewidth]{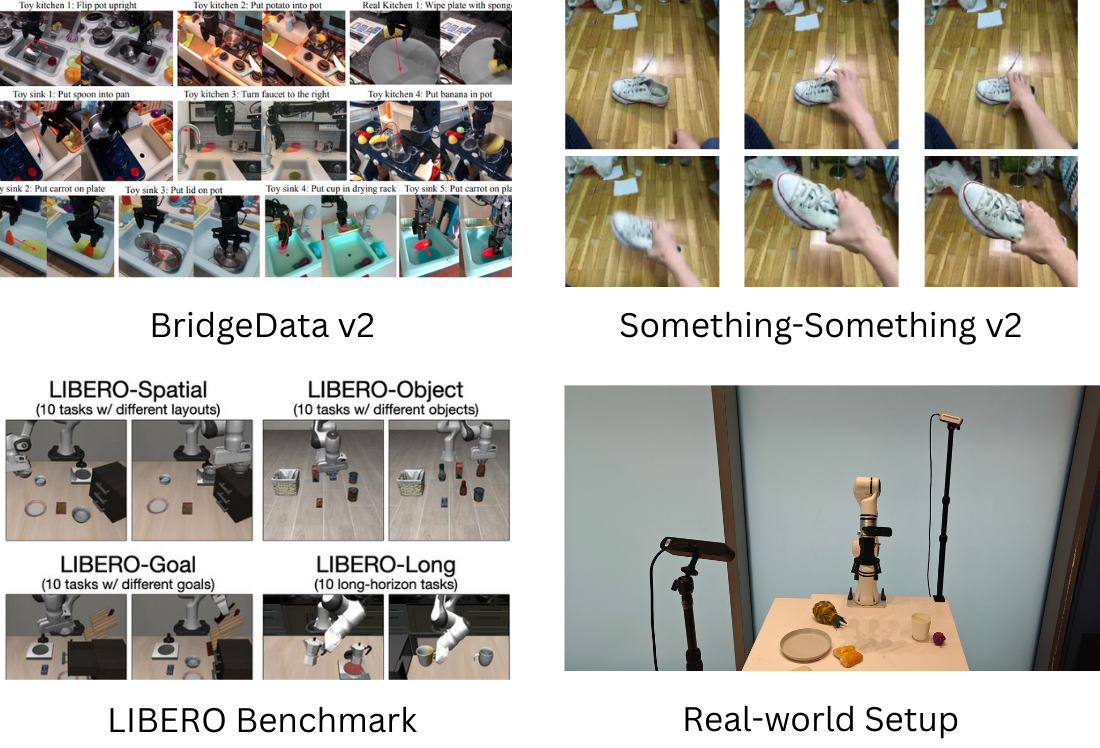}
    \caption{Benchmarks used for pretraining and finetuning.}
    \label{fig:datasets}
\end{figure}

\section{Compute Resources}

All experiments were conducted on a single A100 GPU with default settings, unless otherwise specified: a DreamerV3 world model with 50M parameters configuration and a latent action space of dimension 7, the same dimension as the ground-truth actions. The action chunk size is set to 10 future actions for BAKU experiments and 16 for Diffusion Policy experiments. We only use 10\% of the Something-Something v2 dataset for pretraining. With these settings, pretraining on each of the datasets BridgeData v2 and Something-Something v2 requires approximately 48 hours, while finetuning on LIBERO-90 takes around 24 hours, and finetuning on the LIBERO suites (Spatial, Object, Goal, Long) takes approximately 3 hours each. Finetuning on our custom real-world dataset takes only about 20 minutes.





\ifSubfilesClassLoaded{
    \bibliography{references}
}{}

\end{document}